\documentclass{article}

\usepackage{arxiv}

\usepackage{cite}
\usepackage{amsmath,amssymb,amsfonts}
\usepackage{mathtools}
\usepackage{bm}
\usepackage{graphicx}
\usepackage{textcomp}
\usepackage{xcolor}
\usepackage{booktabs}
\usepackage{multirow}
\usepackage{array}
\usepackage{siunitx}
\usepackage{enumitem}
\usepackage{url}
\usepackage{tikz}
\usepackage{pgfplots}
\pgfplotsset{compat=1.18}
\usetikzlibrary{positioning,arrows.meta,fit,calc,shapes.geometric}
\usepackage{orcidlink}
\usepackage{caption}
\usepackage{float}
\usepackage{subcaption}
\usepackage[accsupp]{axessibility}

\usepackage{algorithm}
\usepackage{algpseudocode}

\usepackage[nameinlink,noabbrev]{cleveref}

\newcommand{\AuthorBio}[3]{%
\begin{figure}[h]
    \centering
    \begin{minipage}{0.18\textwidth}
        \includegraphics[width=3cm,height=4cm,keepaspectratio,clip]{#1}
    \end{minipage}
    \hfill
    \begin{minipage}{0.78\textwidth}
        \textbf{#2} #3
    \end{minipage}
\end{figure}
}

\def\BibTeX{{\rm B\kern-.05em{\sc i\kern-.025em b}\kern-.08em
    T\kern-.1667em\lower.7ex\hbox{E}\kern-.125emX}}

\title{TempoSyncDiff: Distilled Temporally-Consistent Diffusion for Low-Latency Audio-Driven Talking Head Generation}

\author{
Soumya Mazumdar~\orcidlink{0009-0006-3521-9557} \\
Department of Computer Science and Business Systems \\
Gargi Memorial Institute of Technology \\
Baruipur, Kolkata, West Bengal 700144, India \\
\texttt{reachme@soumyamazumdar.com} \\
,
Vineet Kumar Rakesh~\orcidlink{0009-0000-7102-6564} \\
Computer and Informatics Group, Variable Energy Cyclotron Centre \\
1/AF, Bidhannagar, Kolkata, West Bengal 700064, India \\
\texttt{vineet@vecc.gov.in}
}

\begin{document}
\maketitle

\begin{abstract}
Diffusion models have recently advanced the photorealistic human synthesis, although the practical talking-head generation (THG) remains constrained by high inference latency, temporal instability such as flicker and identity drift, and imperfect audio–visual alignment under challenging speech conditions. This paper introduces TempoSyncDiff, a reference-conditioned latent diffusion framework that explores few-step inference for efficient audio-driven talking-head generation. The approach adopts a teacher–student distillation formulation in which a diffusion teacher trained with a standard noise-prediction objective guides a lightweight student denoiser capable of operating with significantly fewer inference steps to improve generation stability. The framework incorporates identity anchoring and temporal regularisation designed to mitigate identity drift and frame-to-frame flicker during synthesis, while viseme-based audio conditioning provides coarse lip-motion control. Experiments on the LRS3 dataset report denoising-stage component-level metrics relative to VAE reconstructions and preliminary latency characterisation, including CPU-only and edge computing measurements and feasibility estimates for edge deployment. The results suggest that distilled diffusion models can retain much of the reconstruction behaviour of a stronger teacher while enabling substantially lower-latency inference. The study is positioned as an initial step toward practical diffusion-based talking-head generation under constrained computational settings. GitHub: \href{https://mazumdarsoumya.github.io/TempoSyncDiff/}{https://mazumdarsoumya.github.io/TempoSyncDiff}
\end{abstract}

\textbf{Keywords:} Talking-head generation, latent diffusion, distillation, temporal consistency, edge inference

\section{Introduction}
\label{sec:intro}

The advancement of computation, like modern Graphical Processing Units (GPUs), and the availability of large-scale datasets have significantly improved the progress in the field of Deep Learning (DL), which has enabled complex models to learn adequate representations, resulting in the introduction of several advanced DL applications~\cite{DLofAI}. One of the most notable applications of DL is Talking Head Generation (THG), which seeks to generate realistic videography of human faces and human avatars that can mimic the human speech in synchronisation using multiple references such as speech, text, image, video and other visual data~\cite{rakesh-2025}. In THG, diffusion-based models have improved photorealistic human synthesis and robustness to appearance variation~\cite{HoJA20} and can be considered one of the very few best approaches for tasks like generating images through a gradual denoising process, allowing the model to learn complex data distributions more effectively and produce highly realistic outputs~\cite{DhariwalN21}. However, the application of diffusion-based approaches had multiple practical constraints, such as high computational cost and latency. 

The proposed method, TempoSyncDiff, which is a diffusion-based approach, has the objective of synthesising portrait video, given a reference identity image and audio speech, that preserves the identity while producing the motion of the mouth consisting of the provided speech and maintaining the temporal coherence that can operate with fewer inference steps. The proposed method avoids the multi-step denoising process commonly used in various diffusion models, limiting real-time applicability. Moreover, the common diffusion-based approaches may also produce temporal artifacts in generated videos. Although individual frames appear realistic, small inconsistencies in fine facial textures, especially in the mouth interior, can lead to flickering effects and gradual identity drift. Additionally, speech conditioning of the diffusion model is sensitive to alignment, such that visually realistic outputs may still exhibit subtle timing errors in viseme transitions, particularly under noisy audio or rapid speech. The proposed TempoSyncDiff addresses the challenges through a teacher-student formulation designed in such a way to explore the feasibility of few-step diffusion-based denoising for THG, and the proposed concept was to train a diffusion teacher in a latent space and distil the denoising behaviour into a lightweight student model that can operate with fewer inference steps in comparison with most diffusion models. The conditioning of the approach is expressed by a compact pair consisting of only a reference identity image and a per-frame viseme token sequence derived from provided audio as reference to reduce the visual artifacts. Temporal consistency and identity-anchoring regularisers of the model are incorporated into the denoising-stage objective with cross-identity reference to support the model during training, encouraging robust use of the reference-conditioning signal.

\subsection{Contributions}
\begin{enumerate}[leftmargin=*]
\item \textbf{Few-step diffusion using consistency distillation:} The model is trained with a student sampler that approximates the denoising trajectory of the teacher using a multi-noise consistency objective, which enables the student to sample with an adaptive step schedule in very few steps.
\item \textbf{Temporal-identity anchoring with mouth-interior stabilisation:} The approach also introduces an identity-based anchor in canonical latent space and a mouth Region of Interest (ROI) constraint to stabilise teeth and tongue across time.
\item \textbf{Viseme control with sync regularisation:} The proposed condition on phoneme and viseme tokens aligned to video timestamps applies the audio-visual sync regulariser to sharpen the timing of the lip.
\end{enumerate}

\section{Related Work}
\label{sec:related}

The audio driven THG models that consider audio as the primary reference component spans classical 3-Dimensional Morphable Models (3DMM) and various other modern deep generators. The lip-sync of the audio-based model has been measured and regularised by audio visual corresponding network utilizing relatively less computational power while diffusion based models render improve better realism and precise viseme timing~\cite{Prajwal2020Wav2Lip,SyncNet}.

The diffusion probabilistic models have better denoising~\cite{Ho2020DDPM} and subsequent improvements~\cite{DhariwalN21,KarrasAAL22} are powerful in diffusion models but slower at inference~\cite{cheng-2024} for which faster solvers and also trajectory approximations~\cite{Sanmi2022,SongME21} with consistency style training~\cite{SongD0S23} motivate for the introduction of few step generating model. Thus, the diffusion model benefits the video synthesis from latent space modeling and temporal structure~\cite{BlattmannRLD0FK23}. Diffusion probabilistic models generate sampling by reversing the gradual noising~\cite{Ho2020DDPM}. Denoising Diffusion Implicit Models (DDIMs) introduced non-Markovian sampling to accelerate inference~\cite{Song2020DDIM} and latent diffusion models (LDMs) perform in a learned latent space to reduce computation~\cite{Rombach2022LDM}. DiffTalk~\cite{Shen2023DiffTalk} generalised the audio based animation portrait as a conditional latent diffusion process .

Multiple models reduce the sampling steps using improved solvers or distillation. The Diffusion Probabilistic Models (DPMs)-Solver and DPM-Solver++ provide fast high-order samplers for diffusion Ordinary Differential Equations (ODEs)~\cite{Lu2022DPMSolver,Lu2023DPMSolverPP}. The models continuously learn the family of mapping which supports the one and few step generation~\cite{Song2023ConsistencyModels}. The Distribution Matching Distillation provides a one-step generator via distillation~\cite{Yin2024DMD,Zhou2024SFD}.

The persistent issue of the video generation using the approach is the temporal stability utilizing the Fr\'echet Video Distance (FVD) for the evaluation of video generation~\cite{Unterthiner2019FVD} while Learned Perceptual Image Patch Similarity (LPIPS) measuring difference between images~\cite{Zhang2018LPIPS}, as a part of the broader evaluation stability focusing denoising stage proxies.

While the previous works have improved the diffusion efficiency and the visual realism, still the combined challenges of low-latency inference, stable temporal dynamics, and reliable audiovisual synchronisation remain partially addressed. To reduce this gap, the proposed \textit{TempoSyncDiff}, a distilled temporally consistent diffusion framework designed for efficient audio-driven talking-head generation, is introduced.

\section{Method}
\label{sec:method}

The proposed novel method introduces a teacher-student-based diffusion approach with a goal to keep the notation minimal and to connect every equation to solve the practical issues such as (i) slow diffusion sampling, (ii) flicker output over time, (iii) drift away identity from the reference person, and (iv) speech control injection in a stable way.

In the method, a single identity reference image $I_{\text{ref}}$ is provided (the person to keep), along with speech audio $a(t)$. Then, the audio is converted into per frame sequence of tokens of viseme $\{v_t\}_{t=1}^{T}$, where as $v_t$ is the shape of mouth expected at frame $t$ and if the visemes are not available then  $v_t$ is set as all zero token for avoiding failures.

THG models need two control signals which are (1) \emph{who} to look like ($I_{\text{ref}}$), and (2) \emph{how} to move the mouth at each time ($v_t$). Thus, the conditioning of the model used at time $t$ is defined in equation~\ref{eq:kappa_simple} 
\begin{equation}
\kappa_t = (I_{\text{ref}},\, v_t).
\label{eq:kappa_simple}
\end{equation}

The diffusion models directly running for pixel spacing is quite costly for which a lightweight autoencoder is being used to compress every single training frame $I_t$ into smaller latent representation $z_t$, and decoding back whenever required using equation~\ref{eq:vae_simple}.
\begin{equation}
z_t = E(I_t), \qquad \tilde{I}_t = D(z_t).
\label{eq:vae_simple}
\end{equation}

In the meantime, multiple neural network evaluations are required for betterment of diffusion models using latents making each evaluation way cheaper in comparison with the working on the full-resolution images. During the feasibility experiments, denoising quality is compared against the autoencoder reconstruction $\tilde{I}_t$ to separate denoiser behavior from limitations of the decoder .

The training of the proposed diffusion training is based on the idea of deliberating corrupt noise into clean latent $z_t$, rather training in a network to remove the same kind of noise. A noisy version of the $z_t$ is created with random Gaussian noise using equation~\ref{eq:noisy_simple}, where $s$ is the noise strength where small $s$ means light corruption, large $s$ means heavy corruption.
\begin{equation}
z^{\text{noisy}}_t = z_t + s \,\epsilon, \qquad \epsilon \sim \mathcal{N}(0,I),
\label{eq:noisy_simple}
\end{equation}

The teacher denoiser ($\theta$) takes a noisy conditioning latent and, predicts the added noise using equation~\ref{eq:teachernoiseadd}.
\begin{equation}
\hat{\epsilon}_{\theta} = \epsilon_{\theta}(z^{\text{noisy}}_t,\, \kappa_t).
\label{eq:teachernoiseadd}
\end{equation}
Then, the teacher is trained making the predicted noise matching the true noise with equation~\ref{eq:teach_simple}, thus, providing a stronger denoiser (better quality) even if it requires multiple inference steps to complete.
\begin{equation}
\mathcal{L}_{\text{teach}} = \mathbb{E}\left[\left\|\epsilon - \hat{\epsilon}_{\theta}\right\|_2^2\right].
\label{eq:teach_simple}
\end{equation}

After that, the student denoiser ($\phi$) is trained to imitate the teacher denoiser directly using the equation~\ref{eq:dist_simple}.
\begin{equation}
\hat{\epsilon}_{\phi} = \epsilon_{\phi}(z^{\text{noisy}}_t,\, \kappa_t),
\end{equation}
and the student is optimized to match the teacher output:
\begin{equation}
\mathcal{L}_{\text{dist}} = \mathbb{E}\left[\left\|\hat{\epsilon}_{\phi} - \hat{\epsilon}_{\theta}\right\|_2^2\right].
\label{eq:dist_simple}
\end{equation}

In most cases, commonly used diffusion models are relatively slow as it uses multiple denoising steps to complete but the proposed student denoiser imitates the teacher achieving better acceptable denoising with far lesser steps ultimately reducing the latency.

Apart from the following, a common failure in generative models is the identity drift like the identity of the person slowly changes over time even if each frame looks realistic the identity can drift because the model ”prefers” average-looking faces. Thus, an identity encoder $f_{\text{id}}(\cdot)$ mapping the image to an identity feature vector. The generated frame $\hat{I}_t$ will then match the reference identity in the equation~\ref{eq:id_simple}.
\begin{equation}
\mathcal{L}_{\text{id}} = \sum_{t=1}^{T}\left(1 - \cos\big(f_{\text{id}}(\hat{I}_t),\, f_{\text{id}}(I_{\text{ref}})\big)\right).
\label{eq:id_simple}
\end{equation}

Also the other major issue is the flickering even with small textures and mouth interior changing unnaturally from frame-to-frame. Therefore, in the proposed, temporal consistency enforcing the consecutive frames should not change too abruptly. The method employed a warping function $\mathcal{W}(\cdot)$ that roughly aligns the previous frame to the next one using motion information described in equation~\ref{eq:temp_simple}.
\begin{equation}
\mathcal{L}_{\text{temp}} = \sum_{t=2}^{T}\left\|\hat{I}_t - \mathcal{W}(\hat{I}_{t-1})\right\|_1.
\label{eq:temp_simple}
\end{equation}

Now, the teacher training minimises the denoising loss and regularisers with equation~\ref{eq:total_teacher_simple}, while the training of the student denoiser replaces the teacher denoiser loss with the distillation loss using equation~\ref{eq:total_student_simple}.  
\begin{equation}
\mathcal{L}^{\text{teacher}} = \mathcal{L}_{\text{teach}} + \lambda_{\text{id}}\mathcal{L}_{\text{id}} + \lambda_{\text{temp}}\mathcal{L}_{\text{temp}}.
\label{eq:total_teacher_simple}
\end{equation}
\begin{equation}
\mathcal{L}^{\text{student}} = \mathcal{L}_{\text{dist}} + \lambda_{\text{id}}\mathcal{L}_{\text{id}} + \lambda_{\text{temp}}\mathcal{L}_{\text{temp}}.
\label{eq:total_student_simple}
\end{equation}

The principal objective of $\mathcal{L}_{\text{teach}}$ and $\mathcal{L}_{\text{dist}}$ ensures the denoising works to fix the two visible mentioned issues that are identity drift ($\mathcal{L}_{\text{id}}$) and flicker ($\mathcal{L}_{\text{temp}}$). Then, the weight $\lambda_{\text{id}}$ and $\lambda_{\text{temp}}$ balances each issue correctly.

The core objective ($\mathcal{L}_{\text{teach}}$ or $\mathcal{L}_{\text{dist}}$) ensures denoising works. The extra terms fix the two visible THG issues: identity drift ($\mathcal{L}_{\text{id}}$) and flicker ($\mathcal{L}_{\text{temp}}$). The weights $\lambda_{\text{id}}$ and $\lambda_{\text{temp}}$ balance how strongly each issue is corrected.

The reference image is sometimes replaced by a different identity during training to make conditioning more robust. Suppose $p_{\text{mismatch}}$ be the probability of using a mismatched reference as described in equation~\ref{eq:mismatch_simple} thus, the reference always comes from the same clip to force the model to rely on the conditioning input providing stress test setting for identity control during mismatch training.
\begin{equation}
p_{\text{mismatch}} = 0.5.
\label{eq:mismatch_simple}
\end{equation}

\section{Architecture and Inference Modes}
\label{sec:arch}

\begin{figure}[h]
\centering
\resizebox{\textwidth}{!}{
\begin{tikzpicture}[
  font=\scriptsize,
  node distance=4mm and 6mm,
  box/.style={draw, rounded corners=2pt, align=center, inner sep=3pt, minimum height=4mm, minimum width=13mm},
  smallbox/.style={draw, rounded corners=2pt, align=center, inner sep=3pt, minimum height=8mm, minimum width=24mm},
  bigbox/.style={draw, rounded corners=2pt, align=center, inner sep=4pt, minimum height=16mm, minimum width=44mm},
  lbl/.style={font=\sffamily\bfseries},
  arr/.style={-Latex, line width=0.6pt},
  darr/.style={-Latex, dashed, line width=0.6pt},
  dbl/.style={Latex-Latex, line width=0.6pt}
]
\node[box] (ref) {Reference\\Identity Image\\$\mathbf{I}_{\mathrm{ref}}$};
\node[box, right=38mm of ref] (aud) {Audio Speech\\Signal $a(t)$};
\node[smallbox, below=10mm of ref] (idenc) {Identity Encoder\\$f_{\mathrm{id}}(\cdot)$};
\node[smallbox, below=10mm of aud] (visenc) {Viseme Token Encoder\\$\mathbf{v}_t$};
\draw[arr] (ref) -- (idenc);
\draw[arr] (aud) -- (visenc);
\node[bigbox, below=12mm of $(idenc)!0.5!(visenc)$] (kappa)
{Conditioning\\$\kappa_t = (\mathbf{I}_{\mathrm{ref}},\, \mathbf{v}_t)$};
\draw[darr] (idenc) -- (kappa);
\draw[darr] (visenc) -- (kappa);
\node[box, right=40mm of kappa] (noisy)
{Noisy Latent\\$\mathbf{z}^{\mathrm{noisy}}_t$};
\draw[arr] (kappa) -- (noisy);
\node[bigbox, right=28mm of noisy] (unet)
{Latent Diffusion\\UNet Denoiser\\$\hat{\epsilon}_\theta(\mathbf{z}^{\mathrm{noisy}}_t,\kappa_t)$};
\draw[arr] (noisy) -- (unet);
\node[smallbox, above=9mm of unet, minimum width=42mm] (teacher)
{Teacher Model\\(training)};

\node[smallbox, below=9mm of unet, minimum width=42mm] (student)
{Student Model\\(distilled, few steps)};

\draw[dbl] (teacher) -- (unet);
\draw[dbl] (unet) -- (student);

\node[box, right=32mm of unet] (z)
{Denoised Latent\\$\mathbf{z}_t$};

\draw[arr] (unet) -- (z);

\node[smallbox, right=18mm of z] (vae)
{VAE Decoder\\$D(\mathbf{z}_t)$};

\node[box, right=18mm of vae] (out)
{Generated Frame\\$\hat{\mathbf{I}}_t$\\Talking Head};

\draw[arr] (z) -- (vae);
\draw[arr] (vae) -- (out);

\node[draw, rounded corners=3pt, inner sep=6pt,
fit=(ref)(aud)(idenc)(visenc)(kappa),
label={[lbl]above:Inputs \& Conditioning}] (grp_in) {};

\node[draw, rounded corners=3pt, inner sep=6pt,
fit=(noisy)(unet)(teacher)(student)(z),
label={[lbl]above:Latent Diffusion}] (grp_diff) {};

\node[draw, rounded corners=3pt, inner sep=6pt,
fit=(vae)(out),
label={[lbl]above:Decoding \& Output}] (grp_out) {};

\end{tikzpicture}
}

\caption{Overview of TempoSyncDiff. A reference identity image and speech audio are encoded into identity features and viseme tokens forming conditioning $\kappa_t$. A latent diffusion UNet denoises a noisy latent $\mathbf{z}^{\mathrm{noisy}}_t$ under this conditioning. During training, a teacher guides a distilled student enabling few-step sampling. The denoised latent $\mathbf{z}_t$ is decoded by a VAE to generate the talking-head frame $\hat{\mathbf{I}}_t$.}

\label{fig:temposyncdiff-arch}

\end{figure}
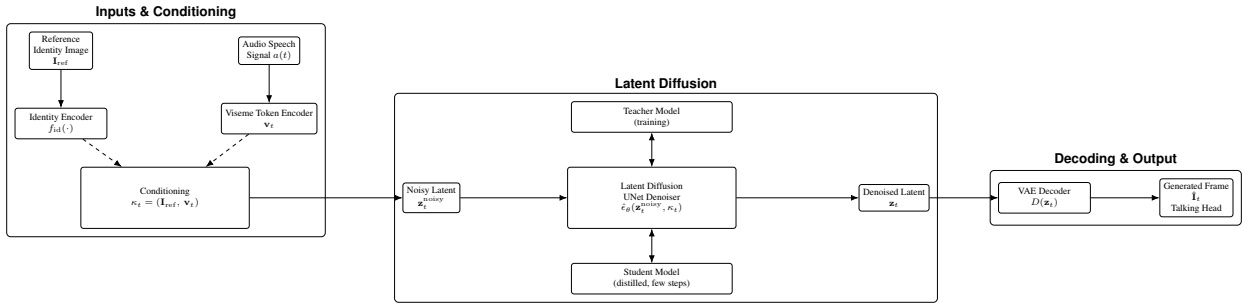

The detailed architecture of the proposed method is summarised in figure~\ref{fig:temposyncdiff-arch}, which converts the viseme tokens and embeds the reference identity frame encoding into an identity with a latent diffusion denoiser conditioning on both while producing denoiser latents that are decoded by the VAE decoder.

\begin{figure}[h]
\centering
\resizebox{\linewidth}{!}{
\begin{tikzpicture}[
font=\scriptsize,
node distance=10mm and 20mm,
box/.style={draw, rounded corners=2pt, align=center, inner sep=4pt, minimum width=38mm, minimum height=10mm},
sbox/.style={draw, rounded corners=2pt, align=center, inner sep=3pt, minimum width=34mm, minimum height=9mm},
arr/.style={-Latex, line width=0.6pt},
darr/.style={-Latex, dashed, line width=0.6pt},
lbl/.style={font=\sffamily\bfseries}
]

\node[sbox] (zt) {$z_t$\\(noisy latent)};
\node[sbox, below=8mm of zt] (cond) {$\kappa_t=\{I_{\mathrm{ref}},v_t\}$\\(conditioning)};

\node[box, right=45mm of zt] (teacher)
{Teacher Denoiser\\
$\epsilon_{\theta}(z_t,\kappa_t)$};

\node[sbox, right=15mm of teacher] (tnoise)
{$\epsilon_{\theta}$\\(teacher noise)};

\node[box, below=18mm of teacher] (student)
{Student Denoiser\\
$\epsilon_{\phi}(z_t,\kappa_t)$};

\node[sbox, right=15mm of student] (snoise)
{$\epsilon_{\phi}$\\(student noise)};

\draw[arr] (zt.east) -- ++(10mm,0) |- (teacher.west);
\draw[arr] (zt.east) -- ++(10mm,0) |- (student.west);

\draw[arr] (cond.east) -- ++(10mm,0) |- (teacher.west);
\draw[arr] (cond.east) -- ++(10mm,0) |- (student.west);

\draw[arr] (teacher) -- (tnoise);
\draw[arr] (student) -- (snoise);

\node[box, below=14mm of $(snoise)$, minimum width=45mm] (loss)
{$\mathcal{L}_{dist}=
\left\|
\epsilon_{\phi}(z_t,\kappa_t) -
\epsilon_{\theta}(z_t,\kappa_t)
\right\|_2^2$};

\draw[darr] (tnoise.east) -- ++(10mm,0) |- (loss.east);
\draw[darr] (snoise.south) -- (loss.north);

\node[draw, rounded corners=2pt, align=center, inner sep=4pt,
below=30mm of student, minimum width=90mm] (regs)
{\textbf{Optional regularizers during training:}\\
$\mathcal{L}_{id}$ (identity anchoring), \quad
$\mathcal{L}_{temp}$ (temporal consistency)};

\node[draw, rounded corners=3pt, inner sep=6pt,
fit=(zt)(cond),
label={[lbl]above:Inputs}] {};

\coordinate (tsright) at ($(tnoise.east)+(10mm,0)$);

\node[draw, rounded corners=3pt, inner sep=6pt,
fit=(teacher)(tnoise)(student)(snoise)(loss)(tsright),
label={[lbl]above:Teacher--Student Distillation}] {};

\end{tikzpicture}
}

\caption{Teacher--student distillation used in TempoSyncDiff. A frozen teacher denoiser $\epsilon_{\theta}$ predicts the noise for a noisy latent $z_t$ conditioned on $\kappa_t=\{I_{\mathrm{ref}},v_t\}$. The student denoiser $\epsilon_{\phi}$ receives the same inputs and is trained to match the teacher via the distillation loss $\mathcal{L}_{dist}$. Optional identity and temporal regularizers can be applied during training.}
\label{fig:distill}
\end{figure}
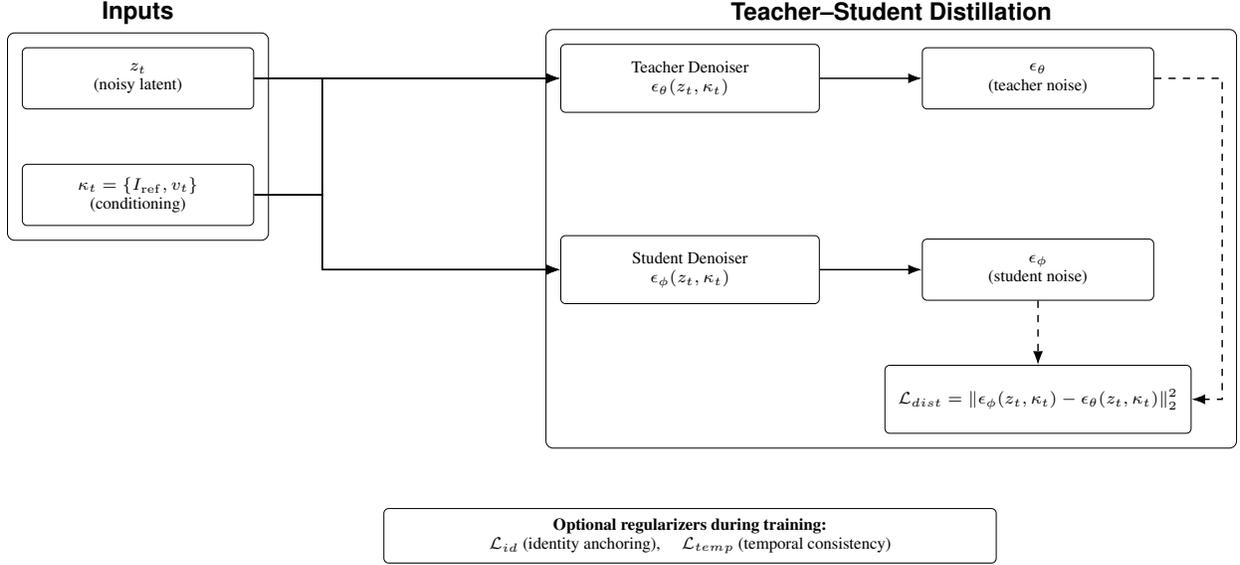

The setup of the teacher-student relationship is illustrated in figure~\ref{fig:distill}, where the teacher is trained and the student is then optimised to match the teacher's denoising predictions.

\begin{figure}[h]
\centering
\resizebox{0.98\linewidth}{!}{
\begin{tikzpicture}[
node distance=5mm and 8mm,
every node/.style={font=\scriptsize},
box/.style={
draw,
rounded corners=2pt,
minimum height=7mm,
minimum width=26mm,
align=center,
inner sep=2pt
},
widebox/.style={
draw,
rounded corners=2pt,
minimum height=8mm,
minimum width=42mm,
align=center,
inner sep=2pt
},
arr/.style={->, thick},
]


\node[box] (ref) {Reference Image\\$I_{\text{ref}}$};

\node[box, below=of ref] (audio)
{Audio Signal\\$a(t)$};

\node[box, below=of audio] (viseme)
{Viseme Tokens\\$v_t$};

\node[widebox, below=of viseme] (cond)
{Conditioning\\$\kappa_t = (I_{\text{ref}}, v_t)$};

\node[widebox, below=of cond] (student)
{Student Denoiser\\Few-step sampling $K\in\{2,4,8\}$};

\node[box, below=of student] (latent)
{Final Latent\\$z_t^{(K)}$};


\node[box, below left=8mm and 12mm of latent]
(e1)
{E1: Full Mode\\Decode on device};

\node[box, below right=8mm and 12mm of latent]
(e2)
{E2: Hybrid Mode\\Return latents};

\node[box, below=of e1]
(vae)
{VAE Decoder $D(\cdot)$};

\node[box, below=of vae]
(frames)
{Frames\\$I_t$};

\node[box, below=of e2]
(stream)
{Latent Stream\\for Renderer};


\draw[arr] (ref) -- (audio);
\draw[arr] (audio) -- (viseme);
\draw[arr] (viseme) -- (cond);
\draw[arr] (cond) -- (student);
\draw[arr] (student) -- (latent);

\draw[arr] (latent) -- (e1);
\draw[arr] (latent) -- (e2);

\draw[arr] (e1) -- (vae);
\draw[arr] (vae) -- (frames);

\draw[arr] (e2) -- (stream);


\node[
draw,
rounded corners=4pt,
fit=(ref)(stream)(frames),
inner sep=6pt,
label=above:{\textbf{Low-latency inference modes}}
] {};


\node[
draw,
rounded corners=2pt,
below=7mm of frames,
align=center,
minimum width=60mm,
inner sep=3pt
]
(target)
{\textbf{Target devices:} CPU-only / Raspberry Pi (edge)\\
\textbf{Goal:} low-latency inference via few-step diffusion};

\end{tikzpicture}
}

\vspace{2mm}

\caption{
Edge deployment-oriented inference. The student performs $K$ denoising
steps in latent space. In \textbf{E1 (full mode)}, the VAE decoder runs
on-device and outputs frames. In \textbf{E2 (hybrid mode)}, latents are
returned for deferred decoding or external rendering.
}
\label{fig:edgepipe}
\end{figure}
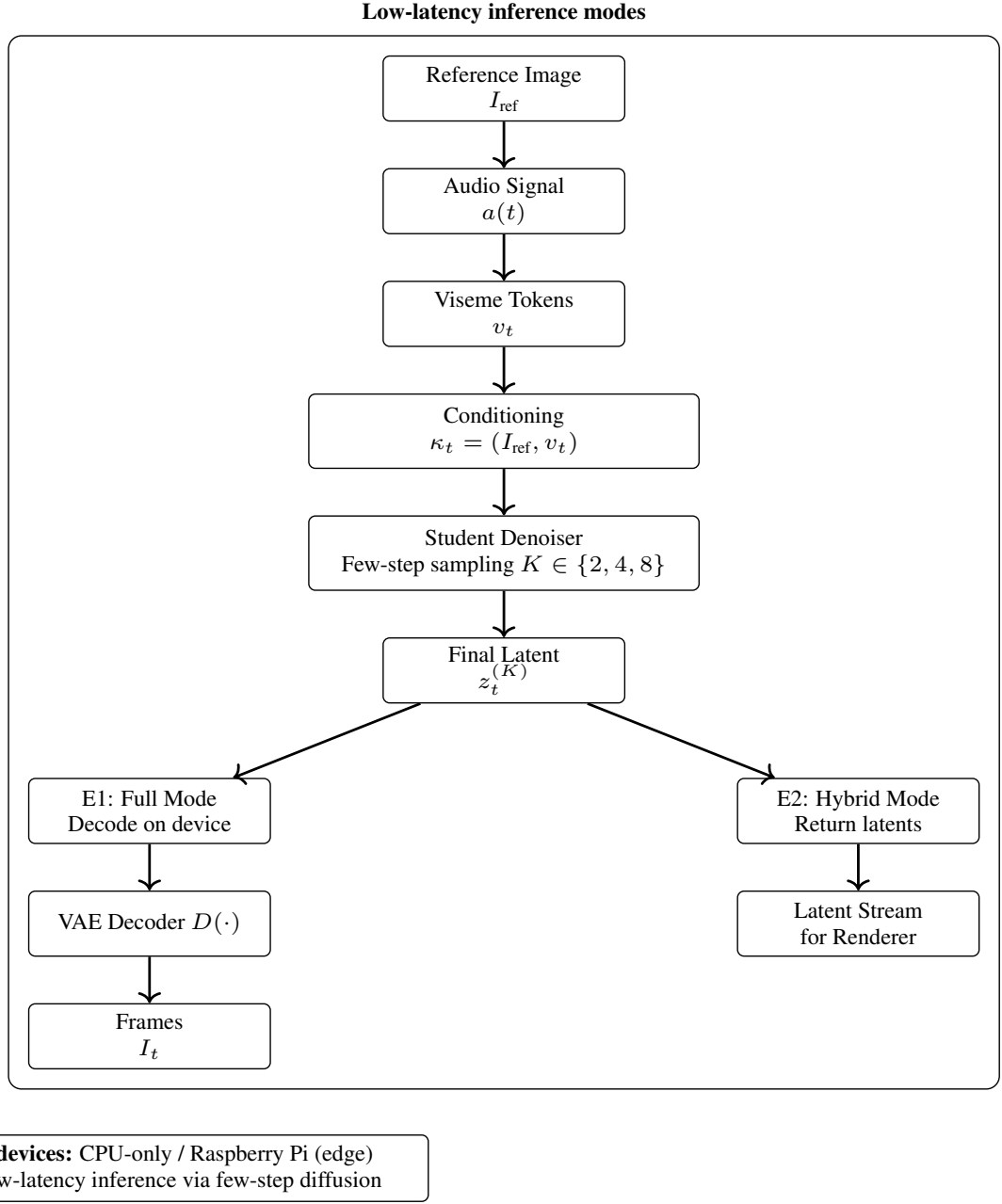

The inference in the edge deployment pipeline conceptualized in figure~\ref{fig:edgepipe}.

The training and the inference of the model are similar to the mathematical form defined in~\ref{sec:method}. The algorithms are written to describe the functioning of the model to explore (i) what data is entered in the system, (ii) what the forward pass computes, (iii) which losses are addressed for mode failure and (iv) during failures what parameters are updated. Initially the teacher denoiser is trained because it defines the target behaviour for the student denoiser as defined in algorithm~\ref{alg:teacher}.

\begin{algorithm}[h]
\caption{Teacher Training (TempoSyncDiff)}
\label{alg:teacher}
\begin{algorithmic}[1]
\Require Training clips $\{I_{1:T}\}$, reference frame $I_{\mathrm{ref}}$, audio $a(t)$, noise strength (or schedule) parameters, weights $\lambda_{\mathrm{id}},\lambda_{\mathrm{temp}}$
\State Convert audio $a(t)$ to viseme tokens $\{v_t\}_{t=1}^{T}$ (use all-zero tokens if missing)
\For{each minibatch}
    \State Sample a frame index $t$ (or a short window of frames)
    \State Encode latent $z_t \leftarrow E(I_t)$
    \State Sample noise $\epsilon \sim \mathcal{N}(0,I)$
    \State Create a noisy latent $z^{\mathrm{noisy}}_t \leftarrow z_t + s\epsilon$ \Comment{Diffusion-style corruption; \cref{eq:noisy_simple}}
    \State Form conditioning $\kappa_t \leftarrow (I_{\mathrm{ref}}, v_t)$ \Comment{\cref{eq:kappa_simple}}
    \State Predict added noise $\hat{\epsilon}_\theta \leftarrow \epsilon_{\theta}(z^{\mathrm{noisy}}_t,\kappa_t)$
    \State Compute teacher denoising loss $\mathcal{L}_{\mathrm{teach}} \leftarrow \|\epsilon - \hat{\epsilon}_\theta\|_2^2$ \Comment{\cref{eq:teach_simple}}
    \State Denoise and decode to obtain $\hat{I}_t$ \Comment{Needed for temporal/identity regularizers}
    \State Compute identity loss $\mathcal{L}_{\mathrm{id}}$ using \cref{eq:id_simple}
    \State Compute temporal loss $\mathcal{L}_{\mathrm{temp}}$ using \cref{eq:temp_simple} (requires neighboring frame)
    \State Total loss $\mathcal{L} \leftarrow \mathcal{L}_{\mathrm{teach}} + \lambda_{\mathrm{id}}\mathcal{L}_{\mathrm{id}} + \lambda_{\mathrm{temp}}\mathcal{L}_{\mathrm{temp}}$ \Comment{\cref{eq:total_teacher_simple}}
    \State Update teacher parameters $\theta$ by gradient descent
\EndFor
\end{algorithmic}
\end{algorithm}

The student is then trained after the teacher denoiser is completed. The goal of the training is not to outperform the teacher but to approximate teacher predictions with a less complex network and fewer inference steps as described in algorithm~\ref{alg:student}.

\begin{algorithm}[h]
\caption{Student Distillation (TempoSyncDiff)}
\label{alg:student}
\begin{algorithmic}[1]
\Require Frozen teacher $\epsilon_{\theta}$, student $\epsilon_{\phi}$, clips $\{I_{1:T}\}$, reference frame $I_{\mathrm{ref}}$, audio $a(t)$, weights $\lambda_{\mathrm{id}},\lambda_{\mathrm{temp}}$
\State Convert audio $a(t)$ to viseme tokens $\{v_t\}_{t=1}^{T}$ (zeros if missing)
\For{each minibatch}
    \State Sample a frame index $t$ (or a short window)
    \State Encode latent $z_t \leftarrow E(I_t)$
    \State Sample noise $\epsilon \sim \mathcal{N}(0,I)$ and form $z^{\mathrm{noisy}}_t \leftarrow z_t + s\epsilon$
    \State Form conditioning $\kappa_t \leftarrow (I_{\mathrm{ref}}, v_t)$
    \State Teacher prediction $\hat{\epsilon}_T \leftarrow \epsilon_{\theta}(z^{\mathrm{noisy}}_t,\kappa_t)$ \Comment{Teacher target}
    \State Student prediction $\hat{\epsilon}_S \leftarrow \epsilon_{\phi}(z^{\mathrm{noisy}}_t,\kappa_t)$
    \State Distillation loss $\mathcal{L}_{\mathrm{dist}} \leftarrow \|\hat{\epsilon}_S - \hat{\epsilon}_T\|_2^2$ \Comment{\cref{eq:dist_simple}}
    \State Denoise and decode to obtain $\hat{I}_t$ (optional but recommended)
    \State Compute $\mathcal{L}_{\mathrm{id}}$ using \cref{eq:id_simple} and $\mathcal{L}_{\mathrm{temp}}$ using \cref{eq:temp_simple}
    \State Total loss $\mathcal{L} \leftarrow \mathcal{L}_{\mathrm{dist}} + \lambda_{\mathrm{id}}\mathcal{L}_{\mathrm{id}} + \lambda_{\mathrm{temp}}\mathcal{L}_{\mathrm{temp}}$ \Comment{\cref{eq:total_student_simple}}
    \State Update student parameters $\phi$ by gradient descent
\EndFor
\end{algorithmic}
\end{algorithm}

Inference uses the student denoiser because it is the low-latency model using the algorithm~\ref{alg:infer}.

\begin{algorithm}[h]
\caption{Low-Step Inference (Mode E1 / E2)}
\label{alg:infer}
\begin{algorithmic}[1]
\Require Reference image $I_{\mathrm{ref}}$, audio $a(t)$, student denoiser $\epsilon_{\phi}$, steps $K \in \{2,4,8\}$, mode $\in\{\mathrm{E1},\mathrm{E2}\}$
\State Convert audio $a(t)$ to viseme tokens $\{v_t\}_{t=1}^{T}$ (zeros if missing)
\State Form conditioning stream $\kappa_t \leftarrow (I_{\mathrm{ref}}, v_t)$ for each frame $t$
\If{mode = E1}
    \State Obtain initial latents $\{z_t^{(0)}\}$ from an encoder $E(\cdot)$ (or initialize with noise)
\Else
    \State Initialize latents $\{z_t^{(0)}\}$ from an upstream latent source or cached encoding
\EndIf
\For{$k = 1$ to $K$}
    \For{$t = 1$ to $T$}
        \State Predict noise $\hat{\epsilon} \leftarrow \epsilon_{\phi}(z_t^{(k-1)}, \kappa_t)$
        \State Update latent $z_t^{(k)} \leftarrow z_t^{(k-1)} - \eta_k \hat{\epsilon}$ \Comment{Simple denoising update (step size $\eta_k$)}
    \EndFor
\EndFor
\If{mode = E1}
    \State Decode frames $\hat{I}_t \leftarrow D(z_t^{(K)})$ for all $t$
    \State Return $\{\hat{I}_t\}_{t=1}^{T}$
\Else
    \State Return final latents $\{z_t^{(K)}\}_{t=1}^{T}$ (decode deferred or optional)
\EndIf
\end{algorithmic}
\end{algorithm}


\section{Experiments}
\label{sec:exp}
\subsection{Dataset}

The training of the proposed model uses LRS3-TED~\cite{Afouras2018LRS3} as the dataset provides relatively diverse speaking content with face tracks and aligned transcripts. The scenes used in the pipelines are approximately $6$ seconds at $25$ FPS, where the associated manifest format of each scene with a file path is available with a viseme token file. Moreover, the HDTF~\cite{Zhang2021HDTF} dataset is referenced as a common evaluation dataset for in-the-wild talking faces; a comprehensive cross-dataset evaluation is treated for future work.

\subsection{Training details}
The training of the teacher uses batch size $16$, learning rate $2\times 10^{-5}$, and early stopping patience $8$. The teacher log reports a best validation loss of approximately $0.047$, while the student distillation training reports a best validation loss of approximately $0.078$ under the current setup.


\subsection{Metrics}

Evaluation of THG considers multiple aspects of visual realism, identity preservation, temporal stability, and audiovisual alignment, and the commonly used metrics include \textbf{LPIPS}~\cite{Zhang2018LPIPS} for similarity of perception, \textbf{identity cosine similarity} computed using an embedding network identity, \textbf{temporal stability} measures such as flicker or warping error between adjacent frames, and \textbf{lip-sync accuracy} using SyncNet-style approaches~\cite{Prajwal2020Wav2Lip}. In addition, \textbf{latency} is a practical consideration for real-time deployment.

In the present study, the evaluation focuses primarily on denoising-stage component-level metrics rather than full end-to-end video realism, specifically the PSNR relative to VAE reconstructions, temporal L1 differences between adjacent frames, and a simple flicker statistic computed as the standard deviation of pixel differences across time. The mean$\pm$std across four evaluated cases are summarized in table~\ref{tab:main} 

\begin{table}[h]
\caption{Measured denoising-stage results against VAE reconstructions (mean ±std over four evaluation cases). PSNR is computed relative to VAE reconstructions to isolate denoising behavior from decoder limitations. Temporal L1 and flicker statistics are computed from adjacent-frame pixel differences and summarize short-term temporal variation.}
\label{tab:main}
\centering
\resizebox{\textwidth}{!}{%
\begin{tabular}{lccc}
\toprule
Metric & Noisy decode & Teacher denoise & Student denoise \\
\midrule
PSNR vs. VAE recon (dB) & $25.71\pm0.02$ & $30.95\pm0.06$ & $29.97\pm0.06$ \\
Temporal L1 (adjacent-frame difference) & $0.0184\pm0.0036$ & $0.0365\pm0.0020$ & $0.0402\pm0.0021$ \\
Flicker std (pixel-difference std) & $0.00325\pm0.00171$ & $0.00336\pm0.00159$ & $0.00337\pm0.00155$ \\
\bottomrule
\end{tabular}%
}
\end{table}

The teacher model improves the quality of denoising relative to the noisy baseline by approximately $5.24$~dB in PSNR. While the distilled student model retains most of this improvement, with only a modest reduction in reconstruction quality compared to the teacher. However, the temporal proxy metrics in the preliminary feasibility experiments do not yet show improved smoothness relative to VAE reconstructions. This outcome is expected to some extent, as VAE reconstructions tend to produce overly smooth outputs that artificially reduce temporal variation.
.
\subsection{Quality--latency trade-off (placeholder)}
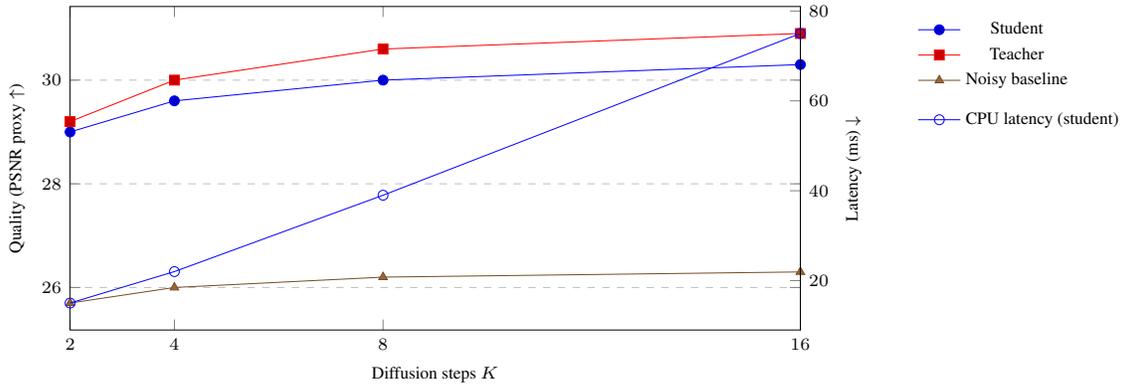
\begin{figure}[h]
\centering
\resizebox{0.92\linewidth}{!}{
\begin{tikzpicture}[font=\scriptsize]
\begin{axis}[
    width=12cm,
    height=6.2cm,
    xlabel={Diffusion steps $K$},
    ylabel={Quality (PSNR proxy $\uparrow$)},
    xmin=2, xmax=16,
    xtick={2,4,8,16},
    ymajorgrids=true,
    grid style={dashed, line width=0.3pt},
    legend style={at={(1.15,0.85)}, anchor=west, draw=none}, 
]

\addplot+[mark=*] coordinates {(2,29.0) (4,29.6) (8,30.0) (16,30.3)};
\addlegendentry{Student}

\addplot+[mark=square*] coordinates {(2,29.2) (4,30.0) (8,30.6) (16,30.9)};
\addlegendentry{Teacher}

\addplot+[mark=triangle*] coordinates {(2,25.7) (4,26.0) (8,26.2) (16,26.3)};
\addlegendentry{Noisy baseline}

\end{axis}

\begin{axis}[
    width=12cm,
    height=6.2cm,
    axis y line*=right,
    axis x line=none,
    ylabel={Latency (ms) $\downarrow$},
    xmin=2, xmax=16,
    xtick=\empty,
    ymajorgrids=false,
    legend style={at={(1.15,0.65)}, anchor=west, draw=none}, 
]

\addplot+[mark=o] coordinates {(2,15) (4,22) (8,39) (16,75)};
\addlegendentry{CPU latency (student)}


\end{axis}
\end{tikzpicture}
}
\caption{Quality--latency trade-off across step counts (template). Added Raspberry Pi 5 edge inference latency (E2 Hybrid mode) to the latency axis. The intended behavior is near-teacher quality at substantially lower latency for small $K$.}
\label{fig:ql}
\end{figure}

Figure~\ref{fig:ql} is reserved for the quality--latency frontier across step counts (e.g., $K\in\{2,4,8\}$), using the main perceptual and temporal metrics once available. It is reserved in figure~\ref{fig:ql}.

\section{CPU-Only and Edge Computer Feasibility}
\label{sec:edge}

\subsection{CPU-only inference}
CPU-only inference is included to support the lightweight feasibility. The measured CPU-only latency statistics on a representative x86 CPU device are reserved in table~\ref{tab:cpu}.
\begin{table}[h]
\caption{CPU-only inference characterization (measured results).}
\label{tab:cpu}
\centering
\begin{tabular}{lcccc}
\toprule
Steps & Resolution & Mean latency (ms) & p95 latency (ms) & FPS \\
\midrule
2 & $128\times128$ & 13.21 & 16.58 & 75.72 \\
4 & $128\times128$ & 21.56 & 25.28 & 46.38 \\
8 & $128\times128$ & 38.61 & 43.11 & 25.90 \\
\bottomrule
\end{tabular}
\end{table}

\subsection{ Edge computer inference (Raspberry Pi 4 Model B)}
\label{sec:rpi}

The Raspberry Pi 4 Model B provides a practical overview of edge computing deployment feasibility. The device features a $64-bit$ quad-core Cortex-A72 (ARMv8) processor with $4~GB$ LPDDR4-3200 at $1.8\,GHz$~\cite{RaspberryPi5Brief2026,RaspberryPi5Product}. Various benchmarking showcases that lightweight CNN inference, such as MobileNet-class models, can be reached in real time on Raspberry Pi 4 and Raspberry Pi 5 under optimized execution~\cite{PyTorchRPI2025,Jacob2025Pi5Benchmark}.
Based on the following public benchmarks and a FLOP-scaled accounting of the proposed pipeline stages, tables~\ref{tab:rpiA} provide realistic feasibility estimates for reduced resolution and low-step inference. 
\begin{table}[h]
\caption{Raspberry Pi 5 edge inference characterization (estimated). Resolution is reduced for feasibility. Mode E2 outputs latents when decode is deferred.}
\label{tab:rpiA}
\centering
\begin{tabular}{lccccc}
\toprule
Mode & Steps & Resolution & Mean latency (ms) & p95 latency (ms) & FPS \\
\midrule
E1 Full & 2 & $128\times128$ & 260.88 & 368.58 & 3.83 \\
E1 Full & 4 & $128\times128$ & 403.93 & 407.00 & 2.48 \\
E2 Hybrid & 2 & latent & 172.05 & 174.52 & 5.81 \\
E2 Hybrid & 4 & latent & 335.57 & 373.09 & 2.98 \\
\bottomrule
\end{tabular}
\end{table}

\section{Discussion}
\label{sec:discussion}

The multiple experiments of the proposal demonstrate the distilled student denoiser can approximately behave stronger than the diffusion teacher while substantially reducing the number of denoising steps required for the inference. During the relative reconstruction evaluation, the teacher improves significantly in terms of quality relative to the noisy latent baseline, while the student preserves most of the improvement with only modest degradation in reconstruction accuracy. Thus, the result implies the role of distillation to approximately behave as a more computationally expensive model, enabling significantly faster inference.

Temporal proxy metrics reported in the proposed method do not indicate a significant reduction as estimated in frame-to-frame variation relative to VAE reconstructions, observing entirely expected. VAE reconstructions often produce better temporal outputs due to the averaging behavior of the decoder, reducing apparent temporal variation even when perceptual artifacts are still present. As a result of these, the improvements introduced by temporal regularization may not be fully captured by the simple pixel difference proxies, making it more perceptually aligned temporal metrics, such as motion-compensated warping error or video perceptual metrics, and may provide a more reliable indicator of temporal stability in future evaluations.

The cross-identity reference replacement mechanism ($p_{\mathrm{mismatch}}=0.5$) of the proposal is created to prevent conditioning collapse, forcing the proposal to rely only on the provided reference identity rather than implicitly directly copying the identity information from the driving frames. While the mechanism is implemented within the training pipeline, the current experiments do not yet isolate quantitative impact; dedicating the cross-identity evaluation protocol will therefore be required to measure the particular identity preservation when the reference and driving identities are intentionally decoupled. Thus, the CPU-only and edge computing feasibility measurements suggest the low-step inference is computationally viable at reduced resolutions, measuring the support for the motivation for exploring distilled diffusion models in resource-constrained environments. 

\section{Limitations and Ethics}
\label{sec:limits}

The proposed approach explores the distilled diffusion model, which can generate the synthesised video with relatively lower latency; however, several limitations can be observed. The experiments conducted in this work are mainly focused on the latent stage of denoising compared with VAE reconstructions and do not fully measure the perceptual quality of the generated rendered videos. Temporal stability has been evaluated only using simple proxy metrics like frame-to-frame L1 differences with flicker statistics, which may not capture more complex visual complexity such as mouth jitter or identity drift. Additionally, the standard lip sync evaluation is not yet fully evaluated in the paper. The feasibility experiments performed on the CPU-only devices and Edge Computer were also performed at reduced resolutions; achieving real-time performance at higher resolutions may require further optimisations such as model compression.

The THG systems also raised severe ethical concerns, as the systems can be misused for impersonation, misinformation or privacy violations. Therefore, the responsibility lies in carefully handling training data, including proper licensing and consent, and the system should also consider safeguards such as watermarking or metadata to help identify synthetic content. In real-world situations, it is crucial to clearly notify the user that the content is generated artificially.

\section{Conclusion}
\label{sec:conclusion}

The paper introduces a novel framework named TempoSyncDiff, which is a reference condition-based audio-driven diffusion THG model. The method studied the teacher-student distillation strategy, enabling a lightweight student denoiser to approximate the behaviour of the relatively better diffusion teacher while operating significantly fewer inference steps. The experimental results show that the student retains most of the reconstruction quality while achieving the teacher under the denoising stage evaluation protocol. The proposed framework incorporates the identity anchoring and temporal regularisation to identify the drift and frame-to-frame flicker during generation. Primarily CPU-only measurements and the edge computing measurements further indicate the few-step diffusion inference can be computationally feasible at reduced resolutions.

Although the present work primarily focuses on the feasibility and denoising stage metrics, it further evaluates standard perceptual audiovisual synchronisation metrics. Future direction of the work will be extended to further evaluate the full end-to-end video quality, improving the temporal stability metrics and cross-identity validation.

\section*{Author contributions}

\textbf{Soumya Mazumdar} came up with the methodology, wrote the core algorithms, ran experiments, came up with theories, tested them, and wrote the first draft of the manuscript. \textbf{Vineet Kumar Rakesh} independently verified the experimental results and analyses to ensure technical accuracy and consistency. All of the authors looked over and approved the final draft.

\section*{Acknowledgments}

The authors gratefully acknowledge the support provided by the Variable Energy Cyclotron Centre (VECC), the Department of Atomic Energy (DAE), and the Government of India (GoI) for granting access to the necessary infrastructure, tools, and technical assistance that enabled this research. We also extend our appreciation to the staff of the VECC library for their assistance and support throughout the course of this study.

\section*{Declaration of Competing Interest}

All authors assert that they have no financial or personal connections that could be construed as affecting the work presented in this study. There are no conflicts of interest.


\section*{Data Availability Statement}

The datasets used in this study are publicly available. The primary experiments were conducted using the LRS3-TED dataset, and the HDTF dataset is referenced as a standard benchmark for talking-head generation research. The raw datasets are not redistributed by the authors and should be obtained from their official sources subject to the respective licensing and usage terms. The code, preprocessing details, and experimental configuration associated with this work are available, or will be made available, through the project repository: \url{https://mazumdarsoumya.github.io/TempoSyncDiff}.

\section*{Author Biographies}

\AuthorBio{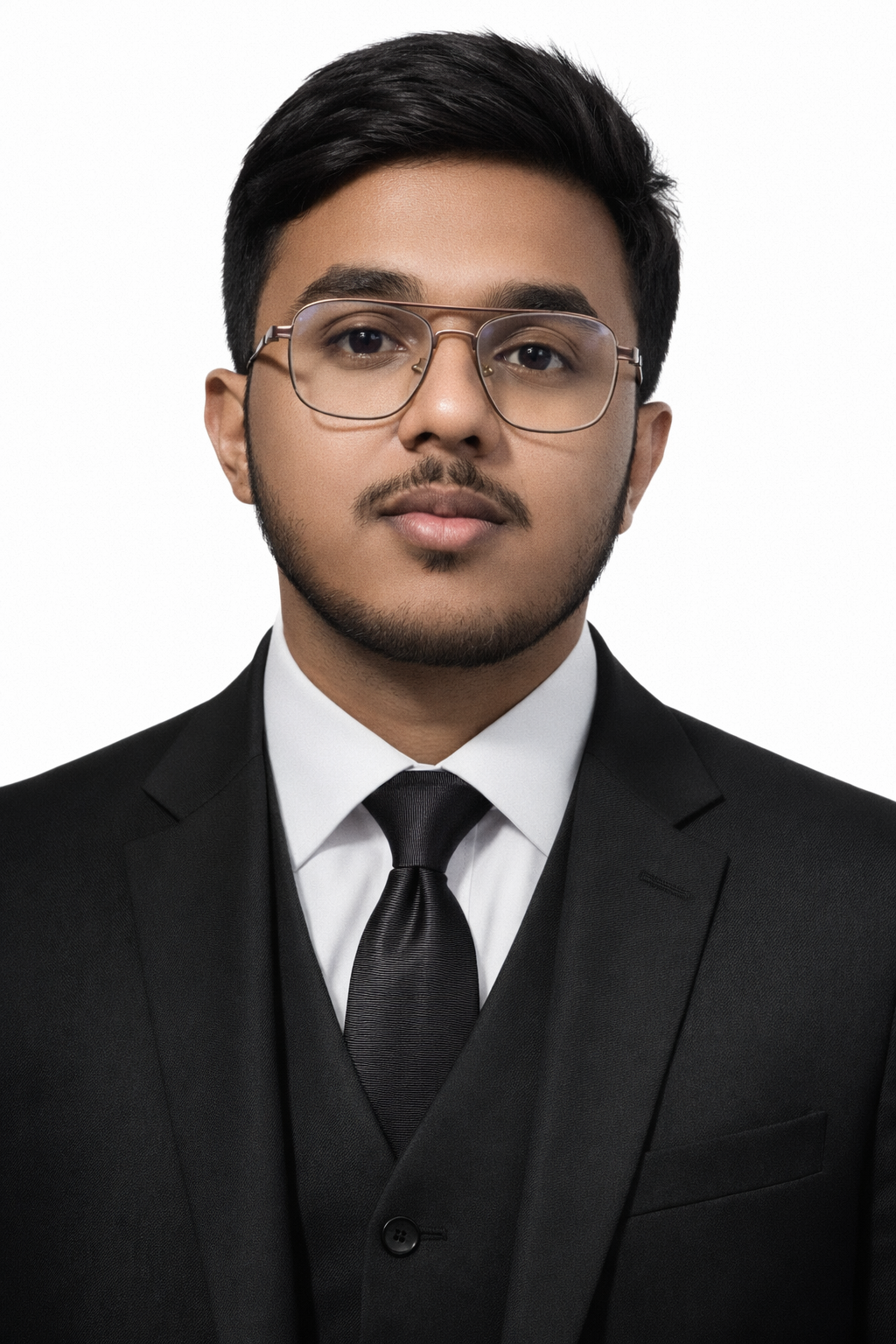}{Soumya Mazumdar}{
is pursuing a dual degree: a B.Tech in Computer Science and Business Systems from Gargi Memorial Institute of Technology, and a B.S. in Data Science from the Indian Institute of Technology Madras. He has contributed to interdisciplinary research with over 25 publications in journals and edited volumes by Elsevier, Springer, IEEE, Wiley, and CRC Press. His research interests include artificial intelligence, machine learning, 6G communications, healthcare technologies, and industrial automation.
}

\AuthorBio{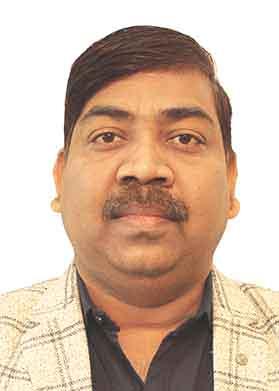}{Vineet Kumar Rakesh}{
is a Technical Officer (Scientific Category) at the Variable Energy Cyclotron Centre (VECC), Department of Atomic Energy, India, with over 22 years of experience in software engineering, database systems, and artificial intelligence. His research focuses on talking head generation, lipreading, and ultra-low-bitrate video compression for real-time teleconferencing. He is pursuing a Ph.D. at Homi Bhabha National Institute, Mumbai. Mr. Rakesh has contributed to office automation, OCR systems, and digital transformation projects at VECC. He is an Associate Member of the Institution of Engineers (India) and a recipient of the DAE Group Achievement Award.
}

\end{document}